%% file: main.tex
\documentclass[sigconf,10pt]{acmart}

\def\BibTeX{{\rm B\kern-.05em{\sc i\kern-.025em b}\kern-.08emT\kern-.1667em\lower.7ex\hbox{E}\kern-.125emX}}

\usepackage{algorithm}
\usepackage{algorithmic}
\usepackage{amsmath}
\usepackage{amssymb}
\usepackage{balance}
\usepackage{bbm}
\usepackage{bmpsize}
\usepackage{booktabs}
\usepackage{cleveref}
\usepackage{enumitem}
\usepackage{epsfig}
\usepackage{graphicx}
\usepackage{multirow}
\usepackage{url}
\usepackage{xcolor}
\usepackage{xifthen}

\newcommand{\stitle}[1]{\smallskip\noindent{\bf #1}}

\newcommand{\prob}[2]{\text{Pr}\big( #1 \ifthenelse{\isempty{#2}}{}{ | #2} \big)}

\newcommand{\set}[1]{\big\{#1\big\}}

\newcommand{\revisenew}[1]{#1}

\DeclareMathOperator*{\argmax}{arg max}
\DeclareMathOperator*{\argmin}{arg min}

\newtheorem{example}{\textbf{Example}}
\newtheorem{theorem}{\textbf{Theorem}}

\copyrightyear{2020} 
\acmYear{2020} 
\setcopyright{acmlicensed}
\acmConference[SIGMOD'20]{Proceedings of the 2020 ACM SIGMOD International Conference on Management of Data}{June 14--19, 2020}{Portland, OR, USA}
\acmBooktitle{Proceedings of the 2020 ACM SIGMOD International Conference on Management of Data (SIGMOD'20), June 14--19, 2020, Portland, OR, USA}
\acmPrice{15.00}
\acmDOI{10.1145/3318464.3380592}
\acmISBN{978-1-4503-6735-6/20/06}

\begin{document}

\title[GOGGLES: Automatic Image Labeling with Affinity Coding]{GOGGLES: Automatic Image Labeling\\ with Affinity Coding}

\input{000-Authors.tex}
\input{001-Abstract.tex}

\maketitle

\input{100-Introduction.tex}
\input{200-Preliminary.tex}
\input{300-GenerateAM.tex}
\input{400-Inference.tex}
\input{500-Experiments.tex}
\input{600-RelatedWork.tex}
\input{700-Conclusion.tex}

\balance

\break
\bibliographystyle{ACM-Reference-Format}
\bibliography{main.bib} 

\end{document}

%% file: 000-Authors.tex
\author{Nilaksh Das}
\affiliation{}
\email{nilakshdas@gatech.edu}

\author{Sanya Chaba}
\affiliation{}
\email{sanyachaba@gatech.edu}

\author{Renzhi Wu}
\affiliation{}
\email{renzhiwu@gatech.edu}

\author{Sakshi Gandhi}
\affiliation{}
\email{sakshi@gatech.edu}

\author{Duen Horng Chau}
\affiliation{}
\email{polo@gatech.edu}

\author{Xu Chu}
\affiliation{}
\email{xu.chu@cc.gatech.edu}

\author{}
\affiliation{%
  \institution{Georgia Institute of Technology}
}

\renewcommand{\shortauthors}{Das, et al.}

%% file: 001-Abstract.tex
\begin{abstract}
Generating large labeled training data is becoming the biggest bottleneck in building and deploying supervised machine learning models. 
Recently, the data programming paradigm has been proposed to reduce the human cost in labeling training data. 
However, data programming relies on designing labeling functions which still requires significant domain expertise. Also, it is prohibitively difficult to write labeling functions for image datasets as it is hard to express domain knowledge using raw features for images (pixels).

We propose \emph{affinity coding}, a new domain-agnostic paradigm for automated training data labeling. 
The core premise of affinity coding is that the affinity scores of instance pairs belonging to the same class on average should be higher than those of pairs belonging to different classes, according to some affinity functions. 
We build the GOGGLES system that implements affinity coding for labeling image datasets by designing a novel set of reusable affinity functions for images, and propose a novel hierarchical generative model for class inference using a small development set.

We compare GOGGLES with existing data programming systems on $5$ image labeling tasks from diverse domains. GOGGLES achieves labeling accuracies ranging from a minimum of $71\%$ to a maximum of $98\%$ without requiring any extensive human annotation. \revisenew{In terms of end-to-end performance, GOGGLES outperforms the state-of-the-art data programming system Snuba by $21\%$ and a state-of-the-art few-shot learning technique by $5\%$, and is  only $7\%$ away from the fully supervised upper bound.}

\end{abstract}

\ccsdesc[500]{Mathematics of computing~Probabilistic inference problems}
\ccsdesc[500]{Computing methodologies~Computer vision representations}
\ccsdesc[300]{Computing methodologies~Cluster analysis}
\ccsdesc[100]{Computing methodologies~Learning settings}

\keywords{affinity coding, probabilistic labels, data programming, weak supervision, computer vision, image labeling}

%% file: 100-Introduction.tex
\section{Introduction}

Machine learning (ML) is being increasingly used by organizations to gain insights from data and to solve a diverse set of important problems, such as fraud detection on structured tabular data, identifying product defects on images, and sentiment analysis on texts. A fundamental necessity for the success of ML algorithms is the existence of sufficient high-quality labeled training data. For example, the current ConvNet revolution would not be possible without big labeled datasets such as the 1M labeled images from ImageNet~\cite{russakovsky2015imagenet}. Modern deep learning methods often need tens of thousands to millions of training examples to reach peak predictive performance~\cite{sun2017revisiting}.
However, for many real-world applications, large hand-labeled training datasets either do not exist, or is extremely expensive to create as manually labeling data usually requires domain experts~\cite{davis2013ctd}.

\noindent \textbf{Existing Work.} We are not the first to recognize the need for addressing the challenges arising from the lack of sufficient training data. The \textbf{ML community} has made significant progress in designing different model training paradigms to cope with limited labeled examples, such as semi-supervised learning techniques~\cite{zhu2005semi}, transfer learning techniques~\cite{pan2010survey} and \revisenew{few-shot learning techniques~\cite{DBLP:journals/corr/XianLSA17,fei2006one,DBLP:journals/corr/abs-1904-05046,2019ChenLKWH19}.  In particular, the most related learning paradigm that shares a similar setup to us, few-shot learning techniques, usually require users to preselect a source dataset or pre-trained model that is in the same domain of the target classification task to achieve best performance. In contrast, our proposal can incorporate as many available sources of information as affinity functions.}

Only recently, the \textit{data programming} paradigm~\cite{ratner2016data} and the Snorkel~\cite{ratner2017snorkel} and Snuba system~\cite{varma2018snuba} that implement the paradigm  were proposed in the \textbf{data management community}. Data programming focuses on reducing the human effort in training data labeling, \revisenew{particularly in unstructured data classification tasks (images, text).}
Instead of asking humans to label each instance, data programming ingests domain knowledge in the form of labeling functions (LFs). Each LF takes an unlabeled instance as input and outputs a label with better-than-random accuracy (or abstain). 
Based on the agreements and disagreements of labels provided by a set of LFs, Snorkel/Snuba then infer the accuracy of different LFs as well as the final probabilistic label for every instance. 
The primary difference between Snorkel and Snuba is that while Snorkel requires human experts to write LFs, Snuba learns a set of LFs using a small set of labeled examples.

While data programming alleviates human efforts significantly, it still requires the construction of a new set of LFs for every new labeling task.
In addition, we find that it is  extremely challenging to design LFs for image labeling tasks primarily because raw pixels values are not informative enough for expressing LFs using either Snorkel or Snuba. After consulting with data programming authors, we confirmed that Snorkel/Snuba require images to have associated metadata, which are either text annotations (e.g., medical notes associated with X-Ray images) or primitives (e.g., bounding boxes for X-Ray images).
These associated text annotations or primitives are usually difficult to come by in practice.

\begin{figure}[t]
\center
\includegraphics[width=0.75\columnwidth]{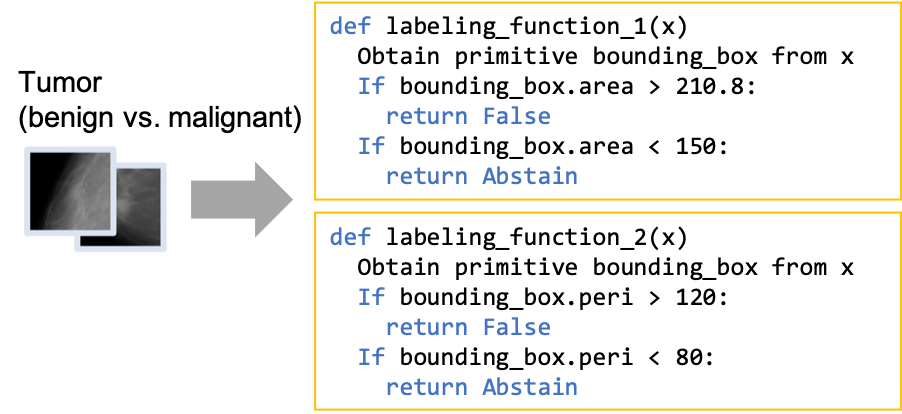}
\vspace{-2mm}
 \caption{Example LFs in data programming~\cite{varma2018snuba}.}
 \label{fig:lfs}
 \vspace{-4mm}
\end{figure}

\begin{example}
\label{ex:intro1}
Figure~\ref{fig:lfs} shows two example labeling functions for labeling an X-Ray image as either benign or malignant~\cite{varma2018snuba}. As we can see, these two functions rely on the bounding box primitive for each image and use the two properties (area and perimeter) of the primitive for labeling. 
We observe that these domain-specific primitives are difficult to obtain. Indeed, ~\cite{varma2018snuba} states, in this particular example, radiologists have pre-extracted these bounding boxes for all images.  
\end{example}

\noindent \textbf{Our Proposal.} We propose \textit{affinity coding}, a new domain-agnostic paradigm for automated training data labeling without requiring any domain specific functions. The core premise of the proposed affinity coding paradigm is that the \textit{affinity scores of instance pairs belonging to the same class on average should be higher than those of instance pairs belonging to different classes, according to some affinity functions}. Note that this is quite a natural assumption --- if two instances belong to the same class, then by definition, they should be similar to each other in some sense.

\vspace{-2mm}
\begin{figure}[h!]
\center
\includegraphics[width=0.85\columnwidth]{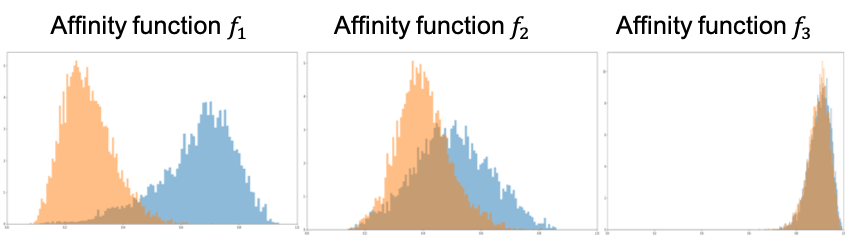}
 \vspace{-4mm}
 \caption{Affinity score distributions. Blue and yellow denote the affinity scores of instance pairs from the same class and different classes, respectively. }
 \label{fig:score_distribution}
 \vspace{-5mm}
\end{figure}

\begin{example}
\label{ex:intro2}
Figure~\ref{fig:score_distribution} shows the affinity score distributions of a real dataset we use in our experiments (CUB) using three of the $50$ affinity functions discussed in Section~\ref{sec:affinity_function}. In this particular case, affinity function $f_1$ is able to distinguish pairs in the same class from pairs in different classes very well; affinity function $f_2$ also has limited power in separating the two cases; and affinity function $f_3$ is not useful at all in separating the classes.
\end{example}

\begin{figure*}[t!]
\center
\includegraphics[width=0.8\linewidth]{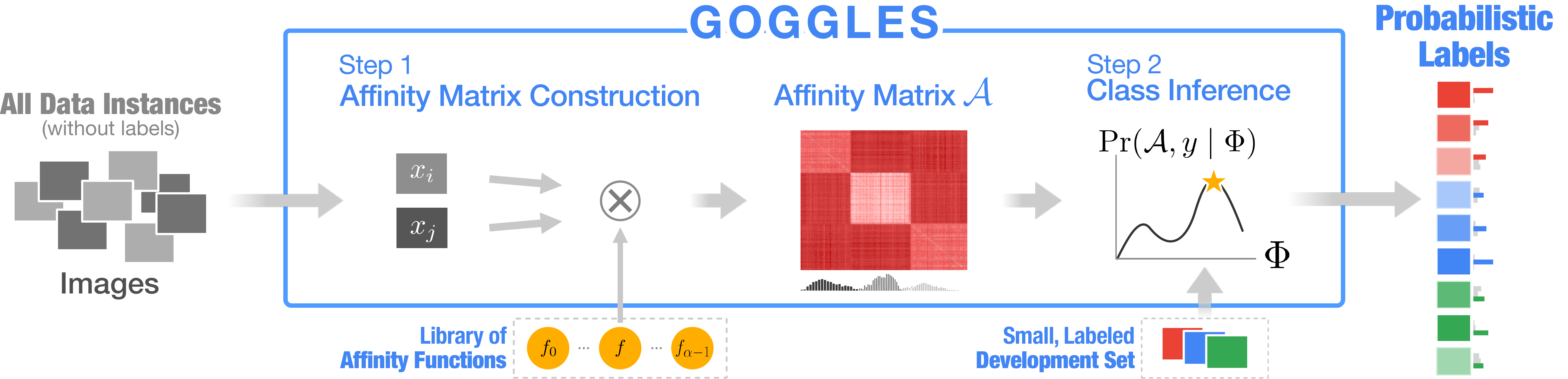}
 \caption{Overview of the GOGGLES framework.}
 \label{fig:framework}
 \vspace{-2mm}
\end{figure*}

We build the GOGGLES system that implements the affinity coding paradigm for labeling image datasets (Figure~\ref{fig:framework}). First, GOGGLES includes a novel set of affinity functions  that can capture various kinds of image affinities. Given a new unlabeled dataset and the set of affinity functions, we construct an affinity matrix. Second, using a very small set of labeled examples (development set), we can assign classes to unlabeled images based on the affinity score distributions we can learn from the affinity matrix. 
Compared with the state-of-the-art data programming systems, our affinity coding system GOGGLES has the following distinctive features. 
\begin{itemize}[leftmargin=*]
    \item Data programming systems need some kinds of metadata (text annotations or domain-specific primitives) associated with each image to express LFs, while GOGGLES makes no such assumptions. 
    \item Assuming the existence of metadata, data programming still requires a new set of LFs for every new dataset. In contrast, GOGGLES is a domain-agnostic system that leverages affinity functions, which are populated once and can be reused for any new dataset.
    \item Both Snorkel/Snuba and GOGGLES can be seen as systems that leverage many sources of weak supervision to infer labels. Intuitively, the more weak supervision sources a system has, the better labeling accuracy a system can potentially achieve. In data programming, the number of sources is the number of LFs. In contrast, affinity coding uses affinity scores between instance pairs under many affinity functions. Therefore, the number of sources GOGGLES has is essentially the number of instances multiplied by the number of affinity functions,  a significantly bigger set of weak supervision sources.
\end{itemize}{}

\vspace{1mm}
\noindent \textbf{Challenges.} We address the following major challenges with GOGGLES:
\begin{itemize}[leftmargin=*]
    \item  The success of affinity coding depends on a set of affinity functions that can capture similarities of images in the same class. However, without knowing which classes and labeling task we may have in the future, we do not even know what are the potential distinctive features for each class. Even if we have the knowledge of the particular distinctive features, they might be spatially located in different regions of images in the same class, which makes it more difficult to design domain-agnostic affinity functions. 
    
    \item Given an affinity matrix constructed using the set of affinity functions, we need to design a robust class inference module that can infer class membership for all unlabeled instances. This is quite challenging for multiple reasons. 
    First, some of the affinity functions are indicative for the current labeling, while many others are just noise, as shown in Example~\ref{ex:intro2}. Our class inference module needs to identify which affinity functions are useful given a labeling task. 
    Second, the affinity matrix is high-dimensional with the number of dimension equals to the number of instances multiplied by the number of affinity functions. In this high-dimensional space, distance between any two rows in the affinity matrix becomes extremely small, and thus making it even more challenging to infer class assignments. 
    Third, while we can infer from the affinity matrix which instances belong to the same class by essentially performing clustering, we still need to figure out which cluster corresponds to which class, relying only on a small development set.
\end{itemize}{}

\stitle{Contributions.} We make the following contributions:
\begin{itemize}[leftmargin=*]
    \item \textbf{The affinity coding paradigm}. We propose \textit{affinity coding}, a new domain-agnostic paradigm for automatic generation of training data. Affinity coding paradigms consists of two main components: a set of affinity functions and a class inference algorithm. 
    To the best of our knowledge, we are the first to propose a domain-agnostic approach for automated training data labeling.

    \item \textbf{Designing affinity functions.} GOGGLES features a novel approach that defines affinity functions based on a pre-trained VGG-16 model \cite{simonyan2014very}. VGG-16 is a commonly used network for representation learning. Our intuition is that  different layers of the VGG-16 network capture different high-level semantic concepts. Each layer may show different activation patterns depending on where a high-level concept is located in an image. 
    We thus leverage all $5$ max-pooling  layers of the network, extracting 10 affinity functions per layer, for a total of $50$ affinity functions. 
    
    \item \textbf{Class inference using hierarchical generative model.}     GOGGLES proposes a novel hierarchical model to identify instances of the same class by maximizing the data likelihood under the generative model. The hierarchical generative model consists of two layers: the base layer consists of multiple Gaussian Mixture Models (GMMs), each modeling an affinity function; and the ensemble layer takes the predictions from each GMM and uses another generative model based on multivariate Bernoulli distribution to obtain the final labels. We show that our hierarchical  generative model  addresses both the curse of dimensionality problem and the affinity function selection problem. We also give theoretical justifications on the size of development set needed to get correct cluster-to-class assignment. 
    \end{itemize}
    
GOGGLES achieves labeling accuracy ranging from a minimum of $71\%$ to a maximum of $98\%$. 
\revisenew{In terms of end-to-end performance, GOGGLES outperforms the state-of-the-art data programming system Snuba by $21\%$ and a state-of-the-art few-shot learning technique by $5\%$, and is  only $7\%$ away from the fully supervised upper bound.}
We also make our implementation of GOGGLES open-source on GitHub\footnote{\url{https://github.com/chu-data-lab/GOGGLES}}.

%% file: 200-Preliminary.tex
\section{Preliminary}
\label{sec:preliminary}
We formally state the problem of automatic training data labeling in Section~\ref{sec:problem_definition}. We then introduce \textit{affinity coding}, a new paradigm for addressing the problem in Section~\ref{sec:affinity_coding_paradigm}.

\subsection{Problem Setup}
\label{sec:problem_definition}
In traditional supervised classification applications, the goal is to learn a classifier $h_\theta$ based on a labeled training set $(x_i, y_i)$, where $x_i \in \mathcal{X}_{train}$ and $y_i \in \mathcal{Y}_{train}$. The classifier is then used to make predictions on a test set.

In our setting, we only have $\mathcal{X}_{train}$ and no $\mathcal{Y}_{train}$. Let $n$ denote the total number of unlabeled data points, and let $y_i^*$ denote the unknown true label for $x_i$.
Our goal is to assign a \textit{probabilistic label} $\Tilde{y}_i^k$ for every $x_i \in \mathcal{X}_{train}$, where $\Tilde{y}_i^k = \prob{[y_i^* = k]}{} \in [0,1]$, where $k \in \set{1, 2, \ldots, K}$ with $K$ being the number of classes in the labeling task, and $\sum_k \Tilde{y}_i^k = 1$.

These probabilistic labels can then be used to train downstream ML models. For example, we can generate a discrete label according to the highest $\Tilde{y}_i^q$ for every instance $x_i$. Another more principled approach is to use the probabilistic labels directly in the loss function $l(h_{\theta}(x_i),y)$, i.e., the expected loss with respect to $\Tilde{y}$: $ \hat{\theta} = \argmin_{\theta}   \sum_{i = 1}^{n} E_{y \sim \Tilde{y}_i}[l(h_{\theta}(x_i),y)]$.
It has been shown that as the amount of unlabeled data increases, the generalization error of the model trained with probabilistic labels will decrease at the same asymptotic rate as supervised models do with additional labeled data~\cite{ratner2016data}.

\subsection{The Affinity Coding Paradigm}
\label{sec:affinity_coding_paradigm}

 We propose \textit{affinity coding}, a domain-agnostic paradigm for automatic labeling of training data. Figure~\ref{fig:framework} depicts an overview of GOGGLES, an implementation of the paradigm.

\noindent \textbf{Step 1: Affinity Matrix Construction.} An affinity function takes two instances and output a real value representing their similarity. 
Given a library of \revisenew{$\alpha$} affinity functions $\mathcal{F} = \{f_0, f_1, \ldots, \revisenew{f_{\alpha-1}}\}$, a set of $n$ unlabeled instances $\{x_0, \ldots, x_{n-1}\}$, and a small $m$ labeled examples $\{(x_{n},y_{n}), \ldots, (x_{n+m-1},y_{n+m-1}) \}$ as the development set, 
we construct an affinity matrix \revisenew{$\mathcal{A} \in \mathbb{R}^{(n+m) \times \alpha(n+m)}$} that encodes all affinity scores between all pairs of instances under all affinity functions. Specifically, the $i^{th}$ row of $\mathcal{A}$ corresponds to instance $x_i$ and every $j^{th}$ column of $\mathcal{A}$ corresponds to the affinity function $f_{j/(n+m)}$ and the instance $x_{j\%(n+m)}$, namely, $\mathcal{A}[i,j] = f_{j/(n+m)}(x_i, x_{j\%(n+m)})$.


\noindent \textbf{Step 2: Class Inference.} 
Given $\mathcal{A}$, we would like to infer the class membership for all unlabeled instances. For every unlabeled instance $x_i, i \in [0,n-1]$, we associate a hidden variable $\Tilde{y}_i$ representing its unknown class. We aim to maximize the data likelihood $\prob{\mathcal{A},\Tilde{y}}{\Phi}$, where $\Phi$ denotes the parameters of the generative model used to generate $\mathcal{A}$.

\stitle{Discussion.} The affinity coding paradigm offers a domain-agnostic paradigm for training data labeling. Our assumption is that, for a new dataset, there exists one or multiple affinity functions in our library $\mathcal{F}$ that can capture some kinds of similarities between instances in the same class. We verify that our assumption holds on all five datasets we tested. It is particularly worth noting that, out of the five datasets, three of them are in completely different domains than the ImageNet dataset the VGG-16 model is trained on. This suggests that our current $\mathcal{F}$ is quite comprehensive. We acknowledge that there certainly exists potential new labeling tasks that our current set of affinity functions $\mathcal{F}$ would fail. 

%% file: 300-GenerateAM.tex
\section{Designing Affinity Functions}
\label{sec:affinity_function}

Our affinity coding paradigm is based on the proposition that examples belonging to the same class should have certain similarities. For image datasets, this proposition translates to \textit{images from the same class would share certain visually discriminative high-level semantic features.}
However, it is nontrivial to design affinity functions that capture these high-level semantic features due to two challenges:
(1) without knowing which classes and labeling task we may have in the future, we do not even know what those features are. 
and (2) even assuming we know the particular features that are useful for a given class, they might be spatially located in different regions of images in the same class.

To address these challenges, GOGGLES leverages pre-trained convolutional neural networks (VGG-16 network~\cite{simonyan2014very} in our current implementation) to transplant the data representation from the raw pixel space to semantic space.
It has been shown that intermediate layers of a trained neural network are able to encode different levels of semantic features, such as edges and corners in initial layers; and textures, objects and complex patterns in final layers \cite{zeiler2014visualizing}.

Algorithm~\ref{alg:vgg} gives the overall procedure of leveraging the VGG-16 network for coding multiple affinity functions. 
Specifically, to address the issue of not knowing which high-level features might be needed in the future, we use different layers of the VGG-16 network to capture different high-level features that might be useful for different future labeling tasks (Line~\ref{line:forlayer}). We call each such high-level feature a \textit{prototype} (Line~\ref{line:all_prototypes}). As not all prototypes are actually informative \revisenew{features}, we keep the \revisenew{top-$Z$} most \revisenew{``activated''} prototypes, which we treat as informative high-level semantic features (Line~\ref{line:top_prototypes}). 
For every one of the informative prototype $v_j^k$ extracted from an image $x_j$, we need to design \revisenew{an} affinity function that checks whether another image $x_i$ has a similar prototype (Line~\ref{line:one_af}). Since these prototypes might be located in different regions, our affinity function is defined to be the maximum similarity between all prototypes of $x_i$ and $v_j^k$ (Line~\ref{line:af_computation}).

We discuss prototype extraction and selection in Section~\ref{sec:prototypes}, and the  computation of affinity functions based on prototypes in Section~\ref{sec:computing_affinity}. 

\begin{algorithm}
\caption{Coding multiple affinity functions $f_0, f_1, \ldots \revisenew{f_{\alpha-1}}$ based on the pre-trained VGG model}
\label{alg:vgg}
\begin{algorithmic}[1]
\REQUIRE Two unlabeled images $x_i$ and $x_j$
\ENSURE Affinity scores between $x_i$ and $x_j$ under $f_0, f_1, \ldots \revisenew{f_{\alpha-1}}$.
\FORALL{each max-pooling layer $L$ in VGG-16} \label{line:forlayer}
   \STATE For image $x_j$, extract all of its prototypes $\boldsymbol{\rho}_j = \set{v_j^{(1,1)}, v_j^{(1,2)}, \ldots, v_j^{H \times W}}$ by passing it through the pre-trained VGG until layer $L$ to obtain a filter map of size $C \times H \times W$, where $C$, $H$ and $W$ are the number of channels, height and width of the filter map respectively and each prototype is vector of length $C$.  \label{line:all_prototypes}
    \STATE Selecting \revisenew{$Z$} most activated prototypes of $x_j$, denoted as $\set{v_j^1, v_j^2, \ldots, \revisenew{v_j^Z}}$ \label{line:top_prototypes}
    \STATE Similarly, for image $x_i$, extract all of its prototypes $\boldsymbol{\rho}_i = \set{v_i^{(1,1)}, v_i^{(1,2)}, \ldots, v_i^{H \times W}}$
    \FORALL {$v_j^k \in \set{v_j^1, v_j^2, \ldots, \revisenew{v_j^Z}}$, where \revisenew{$z \in [1,Z]$}} \label{line:one_af}
    	\STATE $\revisenew{f_{L}^z(x_i, x_j)} \gets \max_{h, w} sim(\revisenew{v_j^Z}, v_i^{(h,w)})$ \label{line:af_computation}
    \ENDFOR
\ENDFOR

\end{algorithmic}
\end{algorithm}
\normalsize

\subsection{Extracting Prototypes}
\label{sec:prototypes}

In this subsection, we discuss (1) how to extract all prototypes from a given image $x_i$ using a particular layer $L$ of the VGG-16 network; and (2) how to select top \revisenew{$Z$} most informative prototypes amongst all the extracted ones.

\noindent \textbf{Extracting all prototypes.} 
To begin, we pass an image $x_i$ through a series of layers until reaching a max-pooling layer $L$ of a CNN to obtain the $F_i = L(x_i)$, known as a \textit{filter map}. We choose max-pooling layers as they condense the previous convolutional operations to provide compact feature representations. 
The filter map $F_i$ has dimensions $C \times H \times W$, where $C$, $H$ and $W$ are the number of channels, height and width of the filter map respectively.
Let us also denote indexes over the height and width dimensions of $F_i$ with $h$ and $w$ respectively.
Each vector $v_i^{(h,w)} \in \mathbb{R}^C$ (spanning the channel axis) in the filter map $F_i$ can be backtracked to a rectangular patch in the input image $x_i$, formally known as the \textit{receptive field} of $v_i^{(h,w)}$.
The location of the corresponding patch of a vector $v_i^{(h,w)}$ can be determined via gradient computation.
Since any change in this patch will induce a change in the vector $v_i^{(h,w)}$, we say that $v_i^{(h,w)}$ encodes the semantic concept present in the patch. Formally, all prototypes we extract for $x_i$ are as follows:
$$\boldsymbol{\rho}_i = \{v_i^{(1,1)}, v_i^{(1,2)}, \ldots, v_i^{(H,W)}\}$$

\begin{figure}[t]
    \center
    \vspace{-4mm}
    \scalebox{0.8}{
    \includegraphics[width=\linewidth]{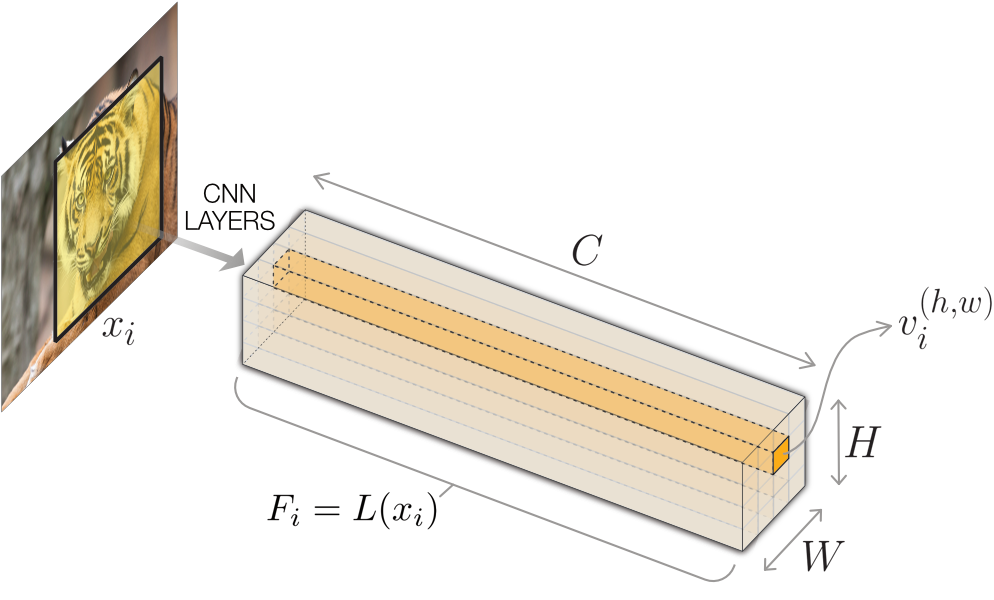}}
    \vspace{-4mm}
    \caption{Extracting all prototypes in a layer of the VGG-16 network. Each prototype corresponds to a patch  in the input image's raw pixel space.}
 \label{fig:filtermap}
 \vspace{-2mm}
\end{figure}

\begin{example}
Figure \ref{fig:filtermap} shows the representation of an image patch in semantic space using a tiger image.  An image $x_i$ is passed through VGG-16 until a max-pooling layer to obtain the filter map $F_i$ that has dimensions $C \times H \times W$. In this particular example, the yellow rectangular patch highlighted in the image is the receptive field of the orange prototype $v_i^{(h,w)}$, which as we can see, captures the ``tiger's head'' concept.
\end{example}

\noindent \textbf{Selecting top \revisenew{$Z$} informative prototypes.} In an image $x_i$, obviously not every patch and the corresponding prototype $v_i^{(h,w)}$ is a good signal. In fact, many patches in an image correspond to background noise that are uninformative for determining its class. Therefore, we need a way to intelligently select the top \revisenew{$Z$} most informative semantic prototypes from all the $H \times W$ possible ones.

In this regard, we first select top-\revisenew{$Z$} channels that have the highest magnitudes of activation. Note that each channel is a matrix $\mathbb{R}^{H \times W}$, and the activation of a channel is defined to be the maximum value of its matrix (typically known as the 2D Global Max Pooling operation in CNNs). We denote the indexes of these top-\revisenew{$Z$} channels as \revisenew{$c_z$}, where \revisenew{$z \in \{1, \ldots, Z\}$}. Based on the top-\revisenew{$Z$} channels, we can thus define the top-\revisenew{$Z$} prototypes as follow:
\begin{align}
    \revisenew{v_{i}^{z}} = v_i^{(h,w)}, \text{where~~} h, w = \argmax_{h, w} F_i[\revisenew{c_z};h;w].
\end{align}
The top-\revisenew{$Z$} prototypes we extract for image $x_i$ are:
$$\boldsymbol{\rho}_i = \{v_{i}^1, v_i^{2}, \ldots, \revisenew{v_i^Z}\}$$
The pair $(h, w)$ may not be unique across the channels, yielding the same concept prototypes. Hence, we drop the duplicate $v_i^{(h,w)}$'s and only keep the unique prototypes.

\begin{example}
We illustrate our approach for selecting top-\revisenew{$Z$} prototypes by an example. Suppose we would like to select top-2 prototypes in a layer that produces the following filter map of dimension $C \times H \times W = 3 \times 2 \times 2$. The three channels are:
\begin{equation}
C_1=
    \begin{bmatrix}
    1       & 0.5  \\
    0.3       & 0.6  \\
\end{bmatrix}
C_2=
    \begin{bmatrix}
    0.1       & 0.7  \\
    0.4       & 0.3  \\
\end{bmatrix}
C_3=
    \begin{bmatrix}
    0.2       & 0.9  \\
    0.5       & 0.1  \\ 
\end{bmatrix}\nonumber
\end{equation}

First, we sort the three channels by the maximum activation in descent order i.e. the maximum element in the matrix: $C_1, C_3, C_2$. Then, we select the first \revisenew{Z}=2 channels: $C_1, C_3$. Next, for each of the selected channels we identify the index of its maximum element on the H and W axis: $(h_1,w_1) = (0,0), (h_2,w_2)=(0,1)$. Finally, we obtain the \revisenew{Z}=2 prototypes by stacking the values over all channels that share the same H and W axis index identified in the last step: \\
$v^1 = \{C_1[h_1,w_1],C_2[h_1,w_1],C_3[h_1,w_1]\}=\{1,0.1,0.2\}$, 
and $v^2 = \{C_1[h_2,w_2],C_2[h_2,w_2],C_3[h_2,w_2]\}=\{0.5,0.7,0.9\}$. 
\end{example}

\subsection{Computing Affinity}
\label{sec:computing_affinity}

Having extracted prototypes for each image, we are ready to define affinity functions and compute affinity scores for a pair of images $(x_i, x_j)$. Affinity functions are supposed to capture various types of similarity between a pair of images. Intuitively, two images are similar if they share some high-level semantic concepts that are captured by our extracted prototypes. Based on this observation, we define multiple affinity functions, each  corresponding to a particular type of semantic concept (prototype). Therefore, the number of affinity functions we can define is equal to the number of max-pooling layers ($5$) of the VGG-16 network multiplied by the number of top-\revisenew{Z} prototypes extracted per layer.

Let us consider a particular prototype \revisenew{$v_j^z$}, that is, the \revisenew{$z^{th}$} most informative prototype of $x_j$ extracted from layer $L$, we define an affinity function as follows:
\begin{align}
\revisenew{f_{L}^z}(x_i,x_j) = \max_{h, w} sim(\revisenew{v_j^z}, v_i^{(h,w)})
\end{align}

As we can see, we calculate the similarity between a prototype \revisenew{$v_j^z$} of $x_j$ and the vector $v_i^{(h,w)} \forall (h,w) \in \set{(1, 1), \dots, (H, W)}$ contained in $F_i = f(x_i)$ using a similarity function $sim(\cdot)$, and pick the highest value as the affinity score.
In other words, our approach tries to find the ``most similar patch" in each image $x_i$ with respect to a given patch corresponding to one of the top-\revisenew{Z} prototypes of image $x_j$. 
We use the cosine similarity metric as the similarity function $sim(\cdot)$ defined over two vectors $a$ and $b$ as follows:
\begin{align}
    sim(a, b) = \frac{a^T b}{\|a\|_2 \|b\|_2}.
\end{align}

\begin{figure}[t]
\center
\includegraphics[width=0.9\columnwidth]{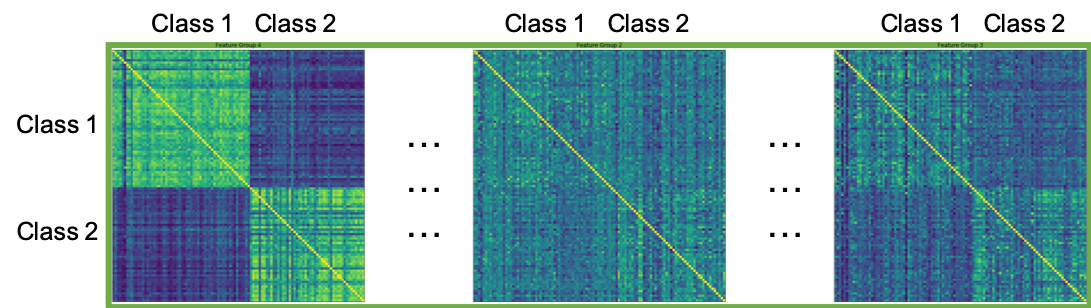}
\vspace{-3mm}
 \caption{An affinity matrix visualized as a heatmap.}
 \label{fig:affinity_matrix}
\vspace{-5mm}
\end{figure}

\begin{example}
\label{ex:example_am}

Figure~\ref{fig:affinity_matrix} shows an example affinity matrix $\mathcal{A}$ for the CUB dataset we use in the experiments. It only shows three of the $50$ affinity functions, which we also used in Example~\ref{ex:intro2}. The rows and columns are sorted by class only for visual intuition. As we can see, some affinity functions are more informative than others in this labeling task. 
 \end{example}

\noindent \textbf{Discussion.} We use all 5 max-pooling layers from the VGG-16. For each max-pooling layer, we use the top-$10$ prototypes, which we empirically find to be sufficient. Note that while we choose VGG-16 to define affinity functions in the current GOGGLES implementation, GOGGLES can be easily extended to use any other representation learning techniques.

In summary, our approach automatically identifies semantically meaningful prototypes from the dataset, and leverages these prototypes for defining affinity functions to produce an affinity matrix.

%% file: 400-Inference.tex
\section{Class Inference}
\label{sec:label_inference}
In this section, we describe GOGGLES' class inference module: given the affinity matrix $\mathcal{A}\in \mathbb{R}^{N \times \revisenew{\alpha}N}$ constructed on $N = n + m$ examples using $s$ affinity functions, where $n$ is the number of unlabeled examples and $m$ is a very small development set (e.g., 10 labeled examples), we would like to assign a class label for every examples $x_i, \forall i \in [1,n]$.
In other words, our goal is to predict $P(y_i = k | \textbf{s}_i)$, where $\textbf{s}_i$ denote the feature vector for $x_i$, namely, the $i^{th}$ row in $\mathcal{A}$, and  $y_i = k \in \{1, 2 \ldots, K\}$ is  a hidden variable representing the class assignment of $x_i$.

\stitle{Generative Modelling of the Inference Problem.} Recall that our main assumption of affinity coding is that the affinity scores of instance pairs belonging to the same class should be different than affinity scores of instance pairs belonging to different classes. In other words, the feature vector $\textbf{s}_i$ of one class should look different than that of another class. This suggests a \textit{generative approach} to model how $\textbf{s}_i$ is generated according to different classes. 
Generative models obtain $P(y_i=k|\textbf{s}_i)$ by invoking the Bayes rules:
\begin{small}
\begin{align}
    \label{eq:bayesRulePosteriorGeneral}
  P(y_i=k|\textbf{s}_i) = \frac{P(y_i=k)P(\textbf{s}_i|y_i=k)}{P(\textbf{s}_i)}  = \frac{\pi_k \times P(\textbf{s}_i|y_i = k)}{\sum_{k'=1}^K \pi_{k'} \times P(\textbf{s}_i|y_i = k')}
\end{align}
\end{small}
where $\pi_k = P(y_i=k)$ is known as the \textit{prior probability} with $\sum_{k'=1}^K \pi_{k'} = 1$, which is the probability that a randomly chosen  instance is in class $k$, and $P(y_i=k|\textbf{s}_i)$ is known as the \textit{posterior probability}. 
To use \Cref{eq:bayesRulePosteriorGeneral} for labeling, we need to learn $\pi_k$ and $P(\textbf{s}_i|y_i = k)$ for every class $k$. $P(\textbf{s}_i|y_i = k)$ is commonly assumed to follow a known distribution family parameterized by \textbf{$\theta_k$}, and is written as $P(\textbf{s}_i| \theta_k)$.
Therefore, the entire set of parameter we need to have to compute \Cref{eq:bayesRulePosteriorGeneral} is $\Theta = \{\pi_1,\dots, \pi_K, \theta_1,\dots, \theta_K\}$. 
A common way to estimate $\Theta$ is by maximizing the log likelihood function:
\begin{small}
\begin{align}
\label{eq:complete_logdatalikelihood}
    L(\mathcal{A}, Y|\Theta) =& \log \prod_{i=1}^{N} P(\textbf{s}_i, y_i|\Theta)  = \sum_{i=1}^{N} \log \big( P(y_i|\Theta)P(\textbf{s}_i|\Theta, y_i) \big)\nonumber \\
         =& \sum_{i=1}^{N} \sum_{k=1}^K \mathbbm{1}_{y_i = k}\log \big( \pi_k P(\textbf{s}_i|\theta_k) \big)\\\nonumber
\end{align}
\end{small}
where $\mathbbm{1}_{.}$ is the identity function that evaluates to 1 if the condition is true and 0 otherwise.
Therefore, the main questions we need to address include (i) what are the generative models to use, namely, the paramterized distributions $P(\textbf{s}_i| \theta_k)$; and (ii) how do we maximize \Cref{eq:complete_logdatalikelihood}.

\stitle{Limitations of Existing Models.} A commonly used distribution is multi-variate Gaussian distribution, where $\theta_k = \{\mu_k, \Sigma_k\}$, where $\mu_k$ is the mean vector and $\Sigma_k$ is the covariance matrix, and $P(\textbf{s}_i| \theta_k)$ is the Gaussian PDF: 
\begin{small}
\begin{equation}
\label{eq:gaussian_pdf}
\begin{aligned}
P(\textbf{s}_i|y_i=k) = P(\textbf{s}_i|\theta_k) = \frac{\exp\{- \frac{1}{2} (\textbf{s}_i - \mu_k)^T \Sigma_k^{-1} (\textbf{s}_i-\mu_k)\}}{(2\pi)^{sn/2}\text{det}(\Sigma_k)^{1/2}} 
\end{aligned}
\end{equation}
\end{small}

This yields the popular Gaussian Mixture Model (GMM), and there are known algorithm for maximizing the likelihood function under GMM. However, a naive invocation of GMM on our affinity matrix $\mathcal{A}$ is problematic:

\begin{itemize}[leftmargin = *]
    \item \textbf{High dimensionality.} The number of feature in the affinity matrix $\mathcal{A}$ is $\revisenew{\alpha}N$. In the naive GMM, the mean vectors and covariance matrices for all classes (components) have \revisenew{$K ({\alpha N \choose 2}+\alpha N)$} number of parameters, which is much larger than the number of examples $N$. It is widely known that the eigenstructure in the estimated covariance matrix $\Sigma_j$ will be significantly and systematically distorted when the number of features exceeds the number of examples~\cite{Dempster1972Mar,Velasco}. 
    
    \item \textbf{Affinity function selection.} Not all affinity functions are useful, \revisenew{as shown in Figure~\ref{fig:affinity_matrix}. If the number of noisy functions is small, GMM naturally handles feature selection as the components will not be well separated by noisy functions and will be well separated by ``good'' functions. However, under such high dimensionality, there could exist too many noisy features that could form false correlations among them and eventually undermine the accuracy of GMM or other generic clustering methods.} 
\end{itemize}

\vspace{-3mm}
\subsection{A Hierarchical Generative Model}
\label{sec:generative}

\begin{figure}[t!]
\center
\scalebox{0.65}{
\includegraphics[width=\columnwidth]{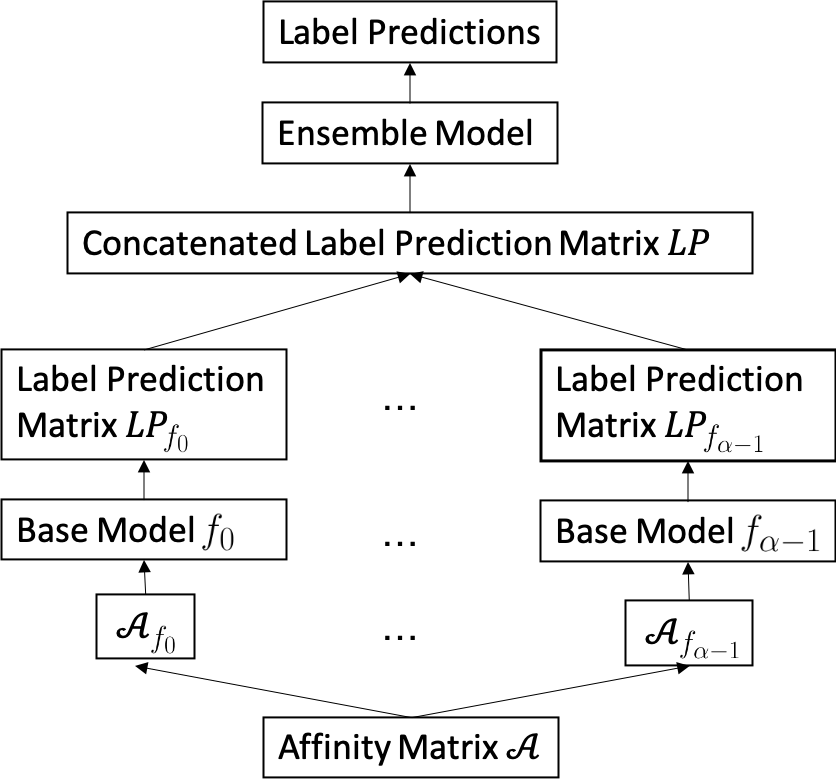}
}
\vspace{-4mm}
 \caption{The hierarchical generative model.}
  \label{fig:hierarchical}
  \vspace{-6mm}
\end{figure}

A fundamental reason for the above two challenges when using GMM is that GMM needs to model  correlations between all pairs of columns, which creates a huge number of parameters and makes it difficult for GMM to determine which affinity functions are more informative. In light of this observation, we propose a \textit{hierarchical generative model} which consists of a set of \textit{base models} and an \textit{ensemble model}, as shown in \Cref{fig:hierarchical}. Each base model captures the correlations of a subset of columns in $\mathcal{A}$ that originate from the same affinity function $f$, and we denote this ``subset'' matrix as $\mathcal{A}_f \in \mathbb{R}^{N \times N}$. The output of each base model is a \textit{label prediction matrix} $LP_f \in \mathbb{R}^{N \times K}$, where the $i^{th}$ row stores the probabilistic class assignments of $x_i$ using affinity function $f$. All label prediction matrices are concatenated together to form the \textit{concatenated label prediction matrix} $LP \in \mathbb{R}^{N \times \revisenew{\alpha}K}$. The ensemble model takes $LP$ and models the correlations of all affinity functions, and produces the final probabilistic labels for each unlabeled instance.

\stitle{The Base Models.} Given the part of the affinity matrix $\mathcal{A}_f \in \mathbb{R}^{N \times N}$ generated by a particular affinity function $f$, the base model aims to predict $P(y_i=k|\textbf{s}_i^f)$, where $\textbf{s}_i^f$ denotes the subset of the feature vector $\textbf{s}_i$ corresponding to $f$.

We design a base generative model for computing $P(y_i=k|\textbf{s}_i^f)$. As discussed before, a generative model requires specifying the class generative distributions $P(\textbf{s}_i^f| \theta_k)$, parameterized by $\theta_k$. We use the popular GMM for this purpose with an important modification. Instead of using the full covariance matrix $\Sigma_k$ that models the correlations between all pairs of columns in $\mathcal{A}_f$, we use the diagonal covariance matrix, which reduces the number of parameters significantly from ${N \choose 2}$ to $N$. Note that this simplification is only possible under the base generative model, as each column of $\mathcal{A}_f$ corresponds to an independent example.

The output of the base model for affinity function $f$ is a label prediction matrix $LP_f \in \mathbb{R}^{N \times K}$, where $LP_f[i,k] = P(y_i=k|\textbf{s}_i^f)$, namely, the probability that affinity function $f$ believes example $x_i$ is in class $k$.

\stitle{The Ensemble Model.} We concatenate  \revisenew{$\alpha$} label prediction matrices $LP_{f_0},\dots, LP_{f_{\revisenew{\alpha}-1}}$ from \revisenew{$\alpha$} base models to obtain the concatenated label prediction matrix $LP \in \mathbb{R}^{N \times \revisenew{\alpha}K}$.  Let $\textbf{s}_i'$ denote the new feature vector of the $i^{th}$ row in $LP$. The goal of the ensemble model is to predict $P(y_i=k|\textbf{s}_i')$.

We again design a generative model for performing the final prediction. As before, we need to decide on a class generative distribution  $P(\textbf{s}_i'|\theta_k')$ parameterized by $\theta_k'$.
The Gaussian distribution used for the base models is not appropriate for the ensemble mode. This is because the values in the concatenated label prediction matrix $LP$ are very close to either $0$ or $1$.
Indeed, in an ideal scenario when all base models work perfectly, all values in $LP$ will be 0 or 1 that correspond to the ground truth. 
This kind of discrete or close to discrete data is problematic for Gaussian distribution which is designed for continuous data. Fitting a Gaussian distribution on this kind of data typically incurs the singularity problem and provides poor predictions~\cite{bishop2006pattern}.
In light of this observation, we convert $LP$ to a one-hot encoded matrix by converting the highest class prediction to $1$ and the rest predictions to $0$ for each instance and each affinity function, and we propose to use a categorical distribution to model $LP$.

After converting $LP$ into a true discrete matrix, Multivariate Bernoulli distribution is a natural fit for modeling $P(\textbf{s}_i'|\theta_k')$, which is parameterized by $\theta_k' =  \{b_{k,1},\dots, b_{k,\revisenew{\alpha}K}\}$:
\begin{equation}
\label{eq:Bernoulli_pdf}
\begin{aligned}
P(\textbf{s}_i'|\theta_k') = \prod_{l=1}^{\revisenew{\alpha}K}b_{k,l}^{\textbf{s}_i'[l]}(1-b_{k,l})^{1-\textbf{s}_i'[l]}
\end{aligned}
\end{equation}

where $\textbf{s}_i'[l]$ is the $l^{th}$ dimension of the binary vector $\textbf{s}_i'$, and we have a total of $\revisenew{\alpha}K$ dimensions. 
The output of the ensemble model is the final label predictions $L \in \mathbb{R}^{N \times K}$, where $L[i,k] = P(y_i=k|\textbf{s}_i')$, namely, the probability that the ensemble model $f$ believes example $x_i$ is in class $k$.

\stitle{Hierarchical Model Address the Two Challenges.} \revisenew{First,} the total number of parameters in the hierarchical model is $2\revisenew{\alpha}KN+\revisenew{\alpha}K$, much smaller than the number of parameters in the naive GMM $K ({\revisenew{\alpha}N \choose 2}+\revisenew{\alpha}N)$, \revisenew{effectively addressing the high-dimensionality problem.} \revisenew{Second,} by consolidating the affinity scores produced by each affinity function to produce a binary $LP$, \revisenew{the ensemble model can only need to model the accuracy of the $\alpha$ affinity functions better instead of the original $\alpha N$ features, and thus can better distinguish the good affinity functions from the bad ones.}

\subsection{Parameter Learning}
\label{sec:em_algorithm}

We need to learn the parameters of the base models and the ensemble model under their respective generative distributions. Expectation-maximization algorithm is the canonical algorithm for maximizing the log likelihood function in the presence of hidden variables~\cite{dempster1977maximum}. We first show the EM algorithm for maximizing the general data log likelihood function in~\Cref{eq:complete_logdatalikelihood}, and then discuss how it needs to modified to learn the base models and the ensemble model.

\stitle{EM for Maximizing~\Cref{eq:complete_logdatalikelihood}} Each iteration of the EM algorithm consists of two steps: an Expectation (E)-step and a Maximization (M)-step. Intuitively, the E-step determines what is the (soft) class assignment $y_i$ for every instance $x_i$ based on the parameter estimates from last iteration $\Theta^{t-1}$. In other words, E-step computes the posterior probability. The M-step takes the new class assignments and re-estimates all parameters $\Theta^{t}$ by maximizing~\Cref{eq:complete_logdatalikelihood}. More precisely, the M-step maximizes the expected value of~\Cref{eq:complete_logdatalikelihood}, since the E-step produces soft assignments.

\begin{enumerate}[leftmargin=*]
 \item \textbf{E Step.} Given the parameter estimates from the previous iteration $\Theta^{t-1}$, compute the posterior probabilities:
  \begin{align}
  \label{eq:e_step}
  \small
    \gamma_{i,k} = P(y_i=k|\textbf{s}_i)=\frac{\pi_k \times P(\textbf{s}_i|\theta_k)}{\sum_{k'=1}^K \pi_{k'} \times P(\textbf{s}_i|\theta_{k'})}
  \end{align}
  \item \textbf{M Step.} Given the new class assignments as defined by $\gamma_{i,k}$, re-estimate $\Theta_t$ by maximizing the following expected log likelihood function:
    \begin{align}
    \label{eq:m_step}
  \mathbb{E}\{L(\mathcal{A}, Y|\Theta) \} =& \sum_{i=1}^N \sum_{k=1}^K\gamma_{i,k}\log \big( \pi_k p(\textbf{s}_i|\theta_k) \big)
  \end{align}
\end{enumerate}

\stitle{EM for Maximizing the Base Model.} For each base model associated with affinity function $f$, $P(\textbf{s}_i|\theta_k)$ in the EM algorithm is replaced with $P(\textbf{s}_i^f|\theta_k^f)$, which is a multivariate Gaussian distribution as shown in \Cref{eq:gaussian_pdf}, but with a diagonal covariance matrix. The entire set of parameters is $\Theta = \{\pi_k^f,\mu_k^f$, $\Sigma_k^f \}$, where $k=0,\dots,K-1$, which update in each M-step as follows:
\begin{small}
\begin{equation}
\label{eq:base_gmm_M_step}
\begin{split}
N_k =& \sum_{i=1}^{N} \gamma_{i,k};  \pi_k^f = N_k/N; \mu_k^f = \frac{1}{N_k}\sum_{i=1}^N \gamma_{i,k}\textbf{s}_i^f\\
    \Sigma_k^f =& \frac{1}{N_k}\sum_{i=1}^N\gamma_{k,i} (\textbf{s}_i^f-\mu_k^f) (\textbf{s}_i^f-\mu_k^f)^T\\
\end{split}
\end{equation}
\end{small}

\stitle{EM for Maximizing the Ensemble Model.} For the ensemble model, $P(\textbf{s}_i|\theta_k)$ in the EM algorithm is replaced with $P(\textbf{s}_i'|\theta_k')$, which is a multivariate Bernoulli distribution, as shown in \Cref{eq:Bernoulli_pdf}. The entire set of parameters is
$\{ \pi_k, b_{k,1}, \\ \dots, b_{k,\revisenew{\alpha}K} \}$, where $k=0,\dots,K-1$, which we update in each M-step as follows:
\begin{small}
\begin{equation}
\label{eq:ensemble_bernulli_M_step}
\begin{split}
N_k =& \sum_{i=1}^{N} \gamma_{i,k};  \pi_k =  N_k/N\\
    b_{k,l} =& \frac{1}{N_k}\sum_{i=1}^N \gamma_{i,k}\textbf{s}_i'[l], \text{~~where~~} l \in \{1, 2, \ldots, \revisenew{\alpha}K\}\\
\end{split}
\end{equation}
\end{small}

\vspace{-3mm}
\subsection{Exploiting Development Set}
\label{sec:exploit_dev_set}

Consider a scenario without any labeled development set, in this case, the hierarchical model essentially clusters all unlabeled examples into $K$ clusters without knowing which cluster corresponds to which class. Following the data programming system~\cite{varma2018snuba}, we assume we have access to a small development set that is typically too small to train any machine learning models, but is powerful enough to determine the correct ``cluster-to-class'' assignment. \revisenew{Note that the theory developed here can also be used to provide theoretical guarantees on the mapping feasibility in the ``cluster-then-label'' category of semi-supervised learning approaches~\cite{zhu2005semi,peikari2018cluster}.}

Let $LS_{k'}$ denote the set of labeled examples for class $k'$. To make our analysis easier, we assume the size of $LS_{k'}$ is the same for all classes. 
Intuitively, we want to map cluster $k$ to class $k'$ if most examples from $LS_{k'}$ are in cluster $k$.  However, this simple cluster-to-class mapping strategy may create conflicting assignments, namely, the same cluster is mapping to multiple classes. 
We propose a more principled way to obtain the one-to-one mapping $g:k \mapsto k'$. We first define the "goodness" of the mapping $L_{g}$ as:
\begin{equation}
\label{eq:mapping_obj_func}
L_{g} = \sum_{k=1}^K \sum_{l \in LS_{g(k)}} \gamma_{l,k}
\end{equation}
To see why $L_{g}$ can represent the "goodness" of a mapping. We represent development sets with a one-hot encoded ground truth matrix $T$ where each element $t_{i,k'}$ is obtained by:
\begin{equation}
    t_{i,k'} = \begin{cases}
        1, \text{ if } i \in LS_{k'}  \\
        0, \text{ otherwise }
    \end{cases}\\
    i = 1,\dots, N;\  k' = 1,\dots, K
\end{equation}
$L_{g}$ is essentially the summation of the element-wise multiplication between the ground truth matrix $T$ and label prediction matrix $LP$ under a column mapping defined by $g$ on the development set. Therefore, $L_{g}$ is expected to be maximized when a mapping $g$ makes the two matrices the most similar under cosine distance.
Given $L_g$, the final mapping $g$ is obtained by:
\begin{equation}
\label{eq:component_class_mapping}
g = \text{arg}\max_{g}\{L_{g}\}, \text{ and } g \text{~ is a one-to-one mapping}
\end{equation}

In other words, the final mapping is a one-to-one mappings that maximizes $L_{g}$. When $K=2$, \Cref{eq:component_class_mapping} becomes
\begin{equation}
\label{eq:component_class_mapping_K_2}
    g(k) =
    \begin{cases}
        k, \text{ if } \sum_{l \in LS_{1}} \gamma_{l,1}  \geq \sum_{l \in LS_{0}} \gamma_{l,1}  \\
        1-k, \text{ otherwise }
    \end{cases}
\end{equation}

\stitle{Algorithm for Solving \Cref{eq:component_class_mapping}.} Instead of enumerating all possible mappings with a complexity of $O(K!)$ (which is actually feasible for a small $K$), we convert it to the \textit{assignment problem} which can be solved in $O(K^3)$.  Let $w_{k,k'}$ denote $w_{k,k'} = \sum{l\in LS_{k'}} \gamma_{l,k}$, 
then \Cref{eq:mapping_obj_func} becomes:
\begin{equation}
\label{eq:assignment_problem}
    L_{g} = \sum_{k=1}^K w_{k,g(k)}
\end{equation}
Finding a $g$ that maximizing \Cref{eq:assignment_problem} is essentially the \textit{Assignment problem}, and there are known algorithms \cite{jonker1987shortest} that solve it with a worst case time complexity of $O(K^3)$.

This ``cluster-to-class'' mapping is performed after obtaining base model predictions and the ensemble model predictions. After the mapping is obtained, we rearrange the the columns in the label prediction matrix $LP_f$ produced by each base model, and the final label matrix $L$ produced by the ensemble according to the mapping $g$, so that the true classes are aligned with the clusters.

\subsection{The Size of Development Set Needed}
\label{ssec:dev_theory}

\begin{figure}[tb]
\centering
\includegraphics[width=0.9\linewidth]{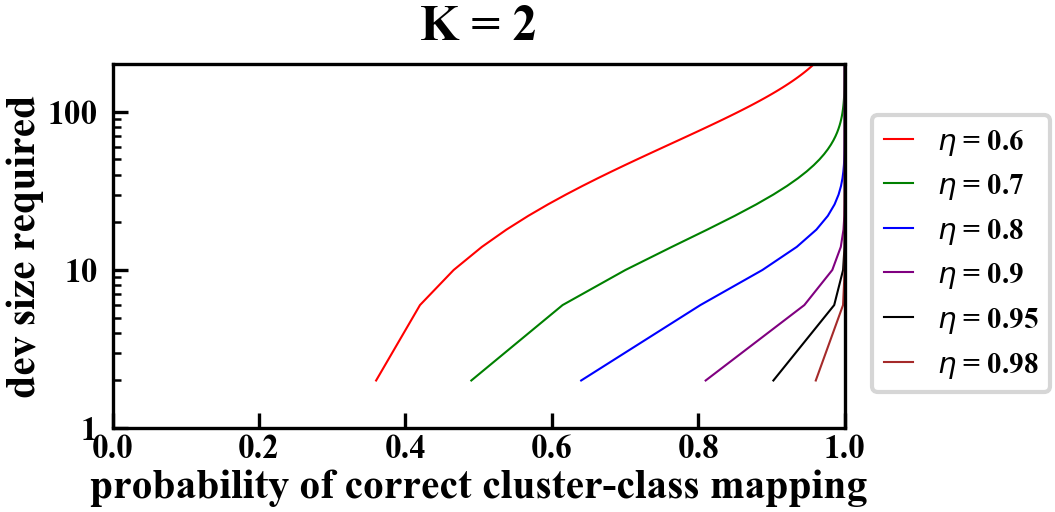}
\vspace{-3mm}
\caption{Size of the Development Set Needed.}
\label{fig:dev_theory_2_classes}
\vspace{-5mm}
\end{figure}

In this section, we give an analysis about the size of the development set needed for GOGGLES to produce the correct ``cluster-to-class'' mapping, where the correct mapping is defined to be the mapping that achieves the highest labeling accuracy, which we denote as $\eta$.  Intuitively, the higher $\eta$ is, the less size we need. Consider an extreme scenario with $K = 2$ classes and our hierarchical generative model produces two clusters that perfectly separate the two classes. In this case, we only need one labeled example to determine which cluster corresponds to which class with $100\%$ confidence. 
\Cref{fig:dev_theory_2_classes} shows the size of development set required when $K=2$ based on our theory to be discussed in the following, we can see when $\eta = 0.8$, only about $20$ examples are required to produces the correct cluster-class mapping with a probability close to 1. However, as we will shown in the experiment section, the number of required development set size is actually much smaller in practice. This is because the theoretical lower-bound we will provide is a rather loose one, for ease of derivation.

\stitle{A Theory on the Size of the Development Set.} let us first assume the mapping of each class is independent,
so the probability of a completely correct mapping is $P_{\text{ind}} = \prod_{k'=1}^K P_{k'}$, 
where $P_{k'}$ denote the probability that class $k'$ is correctly mapped to its corresponding cluster.

To simplify derivation, we further assume "hard" assignment of classes labels: an example is only assigned to one cluster, in other words, $\gamma$ only contains 0 and 1. This is a natural assumption because values in $\gamma$ will be converted to be binary anyway when evaluating the accuracy of the algorithm. In the development set, we have a labeled set $LS_{k'}$ with a size of $d=m/K$ for every class $k'$. Let $d_{k',j},\ j=1,\dots K$, denote the number of examples in the development set $LS_{k'}$ that are in the $j$th cluster, so $\sum_{j=1}^K d_{k',j}= d$. 
Under the independence assumption, \Cref{eq:component_class_mapping} becomes \begin{equation}
\label{eq:component_class_mapping_ind}
    g^{-1}(k') = \text{arg}\max_{1\leq j \leq K} \sum_{i\in LS_{k'}} \gamma_{i,j}
\end{equation}
where $g^{-1}$ denote the inverse mapping of $g$, that is $k' \mapsto k$.
\Cref{eq:component_class_mapping_ind} means that each each class is mapped to the cluster in which its majority lies, so class $k'$ is mapped to its correct cluster only when the majority of $LS_{k'}$ are in that cluster. Assume the $k$th cluster is the correct cluster for class $k'$, so the probability of the $k'$ class mapped correctly is:
\begin{small}
\begin{equation}
\label{eq:prob_correct_mapping}
\begin{aligned}
P_{k'} > Pl_{k'} = & \sum_{d_{k',k} = 0}^{d} P(d_{k',k} > \max_{1\leq j\leq K, j\neq k}{d_{k',j}})\\
=&\sum_{d1,\dots,d_K}P(d_{k',1},\dots, d_{k',K}) \ \textrm{s.t.}\  d_{k',k} > \max_{1\leq j\leq K, j\neq k}d_{k',j}
\end{aligned}
\end{equation}
\end{small}
The first $>$ sign is because on the right side we don't consider the situations with ties in majority vote (the second $>$ sign), where we break the ties randomly and a correct mapping is also possible.
The $P_{\text{ind}}$ is then lower bounded by:
\begin{small}
\begin{equation}
\label{eq:P_ind_lower_bound}
    P_{\text{ind}} > \prod_{k'=1}^K Pl_{k'}
\end{equation}
\end{small}

Suppose the accuracy of our algorithm $\eta$ is known, so the probability of an example being predicted to be its true label equals to $\eta$. An example in the development set $LS_{k'}$ is predicted to be its true label by the algorithm only when it is in the correct cluster, so the probability of it being in the correct cluster equals to $\eta$.
In case of incorrect assignment, we assume the probability of assigning to every possible incorrect classes is equal, being $\rho = \frac{\eta}{K-1}$.
$d_{k',1}, \dots, d_{k',K}$ follow a multinomial distribution:
\begin{small}
\begin{equation}
\label{eq:multinomial}
\begin{aligned}
P(d_{k'1},\dots, d_{k',K}) =& \frac{d!}{\prod_{j=1}^Kd_{k',j}!}\eta^{d_{k',1}}\rho^{d_{k',2}}\rho^{d_{k',3}}\dots \rho^{d_{k',K}}\\
=&\frac{d!}{\prod_{j=1}^Kd_{k',j}!}\eta^{d_{k',k}}\rho^{d-d_{k',k}}
\end{aligned}
\end{equation}
\end{small}
The correct mapping under independent assumption requires the mapping of every class to be correct on their own. This is a rather strict assumption. Without assuming independence, \Cref{eq:component_class_mapping} is able to produce a completely correct mapping when some classes would otherwise fail to be mapped correctly on their own. In other words, the probability of a completely correct mapping is:
\begin{equation}
\label{eq:correct_mapping_geq_ind}
    P_{\text{correct}} \geq P_{\text{ind}}
\end{equation}

Combining \Cref{eq:correct_mapping_geq_ind} and \Cref{eq:P_ind_lower_bound}, we get the following theorem. 

\begin{theorem}
 The probability that \Cref{eq:component_class_mapping} gives the optimal mapping is lower bounded by $P_{\text{correct}} > \prod_{k'=1}^K Pl_{k'}$, where $Pl_{k'}$ is obtained by \Cref{eq:prob_correct_mapping}.

Therefore, the size of development set $m^*$ that produces an optimal mapping with a probability of as least $p$ is given by $m^* = K d^*$, where $d^*$ is the smallest value of $d$ that makes $ \prod_{k'=1}^K Pl_{k'} \geq p$.
\end{theorem}

The time complexity of solving the right hand side in \Cref{eq:prob_correct_mapping} by a brute-force iteration over all combinations of $d_j$ is $O(d!)$, but it can be solved in $O(Kd^2)$ using a dynamic programming based approach.

For ease of of notation, we assume the 1st cluster is the correct cluster for class $k'$. Let $S(j,D_j)$ denote the following:
\begin{small}
\begin{equation}
\begin{aligned}
&S(j,D_j) = \sum_{d_{k',j},...,d_{k',K}} P(d_{k',1},...,d_{k',K})\\
&\textrm{s.t.}\ \ \max_{j\leq l\leq K, l\neq k}d_{k',l} <  d_{k',1} \ \  \text{and}\ \  \sum_{l=j}^K d_{k',l}=D_j
\end{aligned}
\end{equation}
\end{small}
so $Pl_{k'} = S(1,d)$, and for each $j$:
\begin{small}
\begin{equation}
\label{eq:dynamic_solve_P}
S(j,D_j) = \sum_{d_{k',j}=0}^{d} S(j+1,D_j - d_j)
\end{equation}
\end{small}
The time complexity of obtaining $Pl_{k}$ by dynamic programming using \Cref{eq:dynamic_solve_P} is $O(Kd^2)$.

%% file: 500-Experiments.tex
\section{Experiments}
\label{sec:experiments-new}
We conduct extensive experiments to evaluate the accuracy of labels generated by GOGGLES.
Specifically, we focus on the following three dimensions:
\begin{itemize}[leftmargin=*]
    \itemsep0em
    \item \emph{Feasibility and Performance of GOGGLES (Section~\ref{sec:performance_overall})}.
        Is it possible to automatically label image datasets using a domain-agnostic approach? How does GOGGLES compare with existing data programming systems? 
    
    \item \emph{Ablation Study (Section~\ref{sec:ablation_study})}.
        How do the two primary innovations in GOGGLES (namely, affinity matrix construction and class inference) compare against other techniques? 

    \item \emph{Sensitivity Analysis (Section~\ref{sec:sensitivity_analysis})}.
    Is GOGGLES sensitive to the set of affinity functions? What is the size of the development set needed for GOGGLES to correctly determine the correct ``cluster-to-class'' mapping? 
\end{itemize}

\vspace{-3mm}
\subsection{Setup}

\subsubsection{Datasets}

We consider real-world image datasets with varying domains to evaluate the versatility and robustness of GOGGLES. 
Since our approach internally uses a pre-trained VGG-16 model for defining affinity functions, we select datasets which have minimal or no overlap with classes of images from the ImageNet dataset \cite{russakovsky2015imagenet}, on which the VGG-16 model was originally trained. 
Robust performance across these datasets show that GOGGLES is domain-agnostic with respect to the underlying pre-trained model. 
We perform our experiments on the following datasets, which are roughly ordered by domain overlap with ImageNet:
\begin{itemize}[leftmargin=*]
    \item \textbf{CUB}: The Caltech-UCSD Birds-200-2011 dataset \cite{wah2011caltech} comprises of 11,788 images of 200 bird species. The dataset also provides 312 binary image-level attribute annotations that help explain the visual characteristics of the bird in the image, e.g., white head, grey wing etc. We use this metadata information for designing binary labeling functions which are used by a data programming system. To evaluate the task of generating binary labels, we randomly sample 10 class-pairs from the 200 classes in the dataset and report the average performance across these 10 pairs for each experiment. These sampled class-pairs are not present in the ImageNet dataset. However, since ImageNet and CUB contain common images of other bird species, this dataset may have a higher degree of domain overlap with the images that VGG-16 was trained on.
    
    \item \textbf{GTSRB}: The German Traffic Sign Recognition Benchmark dataset \cite{Stallkamp2012} contains 51,839 images for 43 classes of traffic signs. Again, for testing the performance of binary label generation, we sample 10 random class-pairs from the dataset and use them for all the experiments. Although this dataset contains images labeled by specific traffic signs, ImageNet contains a generic ``street sign'' class, and hence this dataset may also have some degree of domain overlap. 
    
    \item \textbf{Surface}: The surface finish dataset \cite{louhichi2019automated} contains 1280 images of industrial metallic parts which are classified as having ``good'' (smooth) or ``bad'' (rough) metallic surface finish. This is a more challenging dataset since the metallic components look very similar to the untrained eye, and has minimal degree of domain overlap with ImageNet.
    
    \item \textbf{TB-Xray}: The Shenzhen Hospital X-ray set \cite{jaeger2013automatic} has 662 images belonging to 2 classes, normal lung X-ray and abnormal X-ray showing various manifestations of tuberculosis. These images are of the medical imaging domain and have absolutely no domain overlap with ImageNet.
    
    \item \textbf{PN-Xray}: The pneumonia chest X-ray dataset \cite{kermany2018identifying} consists of 5,856 chest X-ray images classified by trained radiologists as being normal or showing different types of pneumonia. These images are also of the medical imaging domain and have no domain overlap with ImageNet.
\end{itemize}{}

\vspace{-2mm}
\stitle{Development Set.} GOGGLES uses a small development set to determine the optimal class mapping for a given label assignment, the same assumption in Snuba~\cite{varma2018snuba}. 
By default, we use only 5 label annotations arbitrarily chosen from each class for this. Hence, for the task of generating binary labels, we use a development set having a size of 10 images for all the experiments. We report the performance of GOGGLES on the remaining images from each dataset.

\subsubsection{Data Programming Systems} 

We compare GOGGLES with existing systems: Snorkel~\cite{ratner2017snorkel} and Snuba~\cite{varma2018snuba}.

\stitle{Snorkel} is the first system that implements the data programming paradigm~\cite{ratner2016data}. Snorkel requires humans to design several \textit{labeling functions} that output a noisy label (or abstain) for each instance in the dataset. 
Snorkel then models the high-level interdependencies between the possibly conflicting labeling functions to produce probabilistic labels, which are then used to train an end model. For image datasets, these labeling functions typically work on metadata or extraneous annotations rather than image-based features since it is very hard to hand design functions based on these features.

Since CUB is the only dataset having such metadata available, we report the mean performance of Snorkel on the 10 class-pairs sampled from the dataset by using the attribute annotations as labeling functions. More specifically, we combine CUB's image-level attribute annotations (which describe visual characteristics present in an image, such as white head, grey wing etc.) with the class-level attribute information provided (e.g., class A has white head, class B has grey wing etc.) in order to design labeling functions. Hence, each attribute annotation in the union of the class-specific attributes acts as a labeling function which outputs a binary label corresponding to the class that the attribute belongs to. If an attribute belongs to both classes from the class-pair, the labeling function abstains.
We used the open-source implementation
provided by the authors with our labeling functions for generating the probabilistic labels for the CUB dataset.

\stitle{Snuba} extends Snorkel by further reducing human efforts in writing labeling functions. However, Snuba requires users to provide per-instance primitives for a dataset (c.f. Example~\ref{ex:intro1}), and the system automatically generates a set of labeling functions using a labeled small development set.

Since all 6 datasets do not come with user-provided primitives, to ensure a fair comparison with Snuba, we consulted with Snuba's authors multiple times. They suggested that we use a rich feature representation extracted from  images as their primitives, which would allow Snuba to learn labeling functions. 
As such, we use the \textit{logits layer} of the pre-trained VGG-16 model for this purpose, as it has been well documented in the domain of computer vision that such feature representations encode meaningful higher order semantics for images~\cite{donahue2014decaf, oquab2014learning}. 
For the VGG-16 model trained on ImageNet, this yields us feature vectors having 1000 dimensions for each image. To obtain densely rich primitives which are more tractable for Snuba, we project the logits output onto a feature space of the top-10 principal components of the entire data determined using principal component analysis \cite{wold1987principal}. We use these projected features having 10 dimensions as primitives for Snuba. Empirical testing revealed that providing more components does not change the results significantly. We also use the same development set size for Snuba and GOGGLES. We used the open-source implementation
provided by the authors for learning labeling functions with automatically extracted primitives and for generating the final probabilistic labels.

\subsubsection{Few-shot Learning (FSL)}
\label{sec:fsl}
\revisenew{
Our affinity coding setup which uses 5 development set labels from each class is comparable to the 2-way 5-shot setup for few-shot learning from the computer vision domain. Hence, we compare GOGGLES's end-to-end performance with a recent FSL approach~\cite{2019ChenLKWH19} that achieves state-of-the-art performance on domain adaptation. We use the same development set used by GOGGLES as the few-shot labeled examples for training the FSL model.}

\revisenew{
The original FSL Baseline implementation
uses a model trained on mini-ImageNet for domain adaptation to the CUB dataset, and achieves better performance than other state-of-the-art FSL methods. For a more comparable analysis, we use a VGG-16 model trained on ImageNet, which is the same pre-trained model GOGGLES uses for affinity coding. Note that our adaptation of the FSL Baseline method achieved a much better performance for domain adaptation on CUB than the original results reported in \cite{2019ChenLKWH19}. The FSL models as well as all end models are trained with the Adam optimizer with a learning rate of $10^{-3}$, same as in \cite{2019ChenLKWH19}.}

\subsubsection{Empirical upper-bound (supervised approach).}
\label{sec:upper_bound}
We also would like to compare GOGGLES' performance with an empirical upper-bound, which is obtained via a typical supervised transfer learning approach for image classification. Specifically, we freeze the convolutional layers of the VGG-16 model and only update the weights of the fully connected layers in the VGG-16 architecture while training. We also modify the last fully connected ``logits'' layer of the architecture to our corresponding number of classes.

\subsubsection{\revisenew{Ablation Study:} Other image representation techniques for computing affinity}
\label{sec:other_affinity_function_designs}
GOGGLES computes affinity scores by extracting prototype representations from intermediate layers of a pre-trained model. 
We compare the efficacy of this representation technique with two other typical methods of image representation used in the computer vision domain.  We compare the predictive capacity of each representation technique by constructing an affinity matrix from each candidate feature representation using pair-wise cosine similarity, and then using our class inference approach for labeling.

\stitle{HOG.} We compare with the histogram of oriented gradients HOG descriptor, which is a very popular feature representation technique used for recognition tasks in classical computer vision literature \cite{weinland2010making, yang2012recognizing}. 
The HOG descriptor \cite{dalal2005histograms} represents an image by counting the number of occurrences of gradient orientation in localized portions of the image.

\stitle{Logits.} We also compare with a modern deep learning-based approach, leveraged by recent works in computer vision \cite{sharif2014cnn, akilan2017late}, that uses an intermediate output from a convolutional neural network as an image's feature representation.  %
We use the logits layer from the trained VGG-16 model in our comparison,
which is the output of the last fully connected layer, before it is fed to the softmax operation.

\subsubsection{\revisenew{Ablation Study:} Baseline methods for class inference}
\label{sec:other_class_inference_methods}
The class inference method in GOGGLES  consists of a clustering step followed by class mapping. We compare our proposed hierarchical  model for clustering with other baseline methods, including \textbf{K-means} clustering, Gaussian mixture modeling with expectation maximization (\textbf{GMM}) and spectral co-clustering (\textbf{Spectral}). Since these clustering methods are incognizant of the structural semantics of our affinity-based features which are derived from multiple affinity functions, we simply concatenate all affinity functions to create the feature set for each dataset, and then feed these features to the baseline methods. As we would like to see the absolute best performance of the baseline clustering approaches, we use the optimal ``cluster-class'' mapping for all baselines.

\subsubsection{Evaluation Metrics.}
\revisenew{We use the train/test split as originally defined in each dataset. We report the \textit{labeling accuracy} on the training set for comparing different data labeling systems, Snorkel, Snuba, and GOGGLES. We follow the same approach used in~\cite{ratner2017snorkel,varma2018snuba} to train an end discriminative model  by using the probabilistic labels generated from each data labeling system as training data and report the \textit{end-to-end accuracy} as the end model's performance on the held-out test set. For labeling tasks, all experiments, including baselines, are conducted 10 times, and we report the average.
}

\input{510-LabelingResults.tex}
\vspace{-2mm}
\subsection{Feasibility and Performance }
\label{sec:performance_overall}

Table~\ref{tab:final-results} shows the end-to-end system labeling accuracy for GOGGLES, Snorkel, Snuba, and a supervised approach that serves as an upper bound reference for comparison. 
(1) GOGGLES achieves labeling accuracies ranging from a minimum of $71\%$ to a maximum of $98\%$.  GOGGLES shows an average of $21\%$ improvement over the state-of-the-art data programming system Snuba.
(2) To ensure a fair comparison, we consulted with authors of Snuba and took their suggested approach of automatically extracting the required primitives. As we can see,  the performances of Snuba on all datasets are just slightly better than random guessing. This is primarily because Snuba is really designed to operate on human annotated primitives (c.f. Example~\ref{ex:intro1}). \revisenew{Furthermore, Snuba's performance degrades if the size of the development set is not sufficiently high. Our experiments showed that indeed, if we increase the development set size for Snuba from 10 to 100 (10x increase) for the PN-Xray dataset, the performance jumps from $55.50\%$ to $67.84\%$. In comparison, GOGGLES still performs better with a development set size of only 10 images.}
(3) We can only use Snorkel on  CUB, as CUB is the only dataset that comes with  annotations that we can leverage as labeling functions. These labeling functions may be considered perfect in terms of coverage and accuracy since they are essentially human annotations. GOGGLES uses minimal human supervision and still outperforms Snorkel on CUB.

\vspace{-2mm}
\subsection{Ablation Study}
\label{sec:ablation_study}
We conduct a series of experiments to understand the goodness of different components in GOGGLES, including the proposed affinity functions and the proposed class inference method. Results are shown in Table~\ref{tab:final-results}.

\stitle{Goodness of Proposed Affinity Functions.}
We compare GOGGLES affinity functions with the two common methods of obtaining the distance between two images: HOG and Logits. We use the two baseline methods to generate affinity matrices and run GOGGLES' inference module on them. GOGGLES' affinity functions outperform the other two on almost all datasets. This is because GOGGLES's affinity functions covers features at different scales and locations.

\stitle{Goodness of Proposed Class Inference.}
We compare GOGGLES' inference module with three representative clustering methods: K-means, GMM, and Spectral co-clustering. All methods use the GOGGLES affinity matrix as input data. Note that the three clustering methods are not able to map the clusters to the classes automatically.
As we would like to see the absolute best performance of the baseline clustering approaches, we use the optimal ``cluster-class'' mapping for all baselines. 
GOGGLES's inference module has the best average performance.
\revisenew{The primary reason for our improvement over generic clustering methods is that our generative model adapts to the design of our affinity matrix. Specifically, our generative model is better at (1) handling the high-dimensionality through using the hierarchical structure and reducing the parameters in the base model by using diagonal covariance matrices; and (2) selecting affinity functions through the ensemble model (c.f. Section~\ref{sec:generative}).} 

\revisenew{In terms of running time, without parallelization, our generative model is $\alpha$ (the number of base models) slower than the GMM model (the best baseline method). However, in practice (and in our experiments), we can parallelize all of the base models using different slices of the affinity matrix.}

\vspace{-2mm}
\subsection{Sensitivity Analysis}
\label{sec:sensitivity_analysis}

\begin{figure}[tb]
\centering
\includegraphics[width=0.8\linewidth]{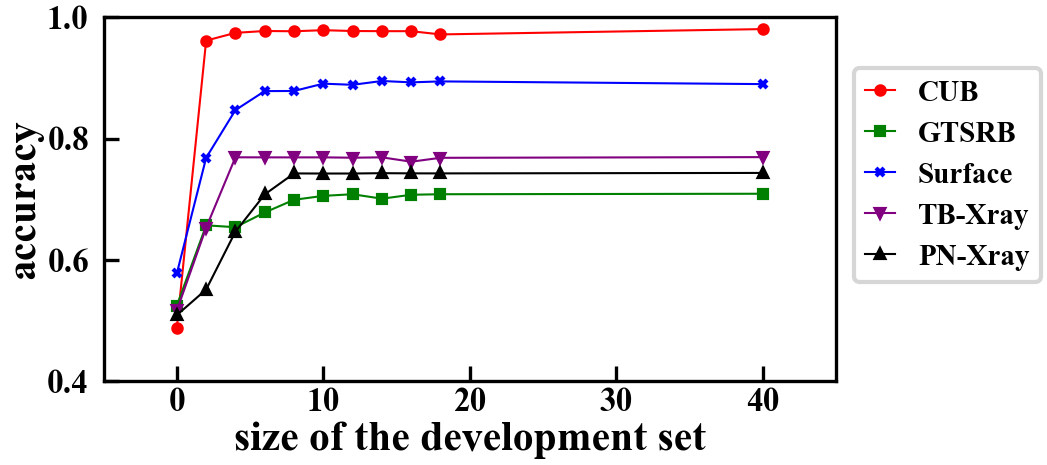}
\vspace{-4mm}
\caption{Accuracy w.r.t. development set size.}
\label{fig:varying_dev_size}
\vspace{-5mm}
\end{figure}

\stitle{Varying Size of the Development Set.}
We vary the size of the development set from 0 to 40 to understand how it affects performance (\Cref{fig:varying_dev_size}). As the development set size increases, the accuracy increases initially, but finally converges. This is expected as when the development set is small, the mapping obtained by \Cref{eq:component_class_mapping} has a low probability being the optimal as predicted in \Cref{fig:dev_theory_2_classes}. When the development set size is large enough, the mapping given by \Cref{eq:component_class_mapping} converges to the optimal mapping, so the accuracy converges.
Another observation is that datasets with higher accuracy converge at a smaller development set size. For example, the CUB dataset has an accuracy of 97.63\% and its accuracy converges at a development set size of 2, while the GTSRB dataset requires a development set size of 8 to converge as it achieves an lower accuracy of 70.75\%. Finally, the empirical size of the development set required to converge is much smaller than the theory predicted in \Cref{fig:dev_theory_2_classes}. A development set with 5 examples per class enough for all datasets.

\begin{figure}[tb]
\centering
\includegraphics[width=0.8\linewidth]{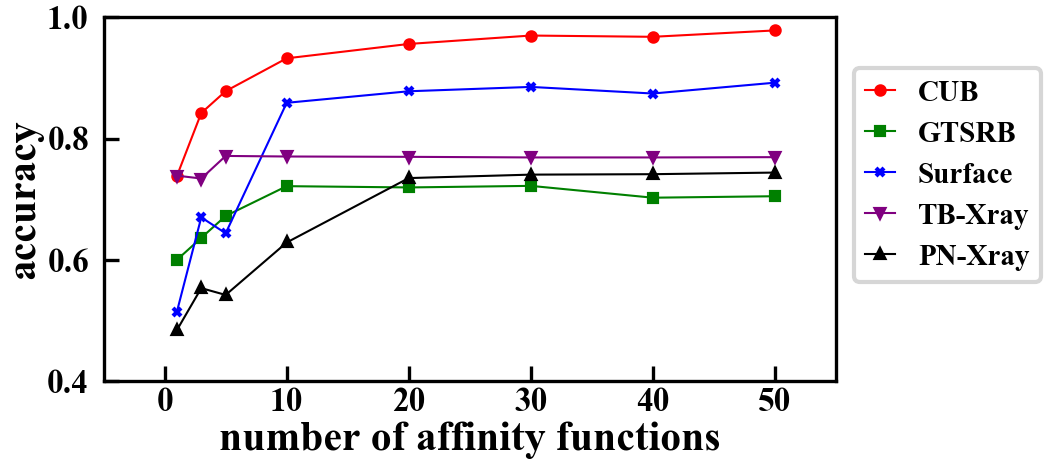}
\vspace{-4mm}
\caption{Accuracy w.r.t. varying \# affinity functions.}
\label{fig:varying_num_AF}
\vspace{-6mm}
\end{figure}

\stitle{Varying Number of Affinity Functions.}
We vary the number of affinity functions to study its affects on the results (\Cref{fig:varying_num_AF}). Accuracy increases as the number of affinity functions increases for all datasets. This is understandable as more affinity functions brings more information that the inference module can exploit.

\vspace{-2mm}
\subsection{End-to-End Performance Comparison.}

\revisenew{
We also use the probabilistic labels generated by Snorkel, Snuba and GOGGLES to train downstream discriminative models following the similar approach taken in~\cite{ratner2017snorkel,varma2018snuba}. Specifically, we use the VGG-16 as the downstream ML model architecture, and tune the weights of the last fully connected layers using cross-entropy loss.
For training the FSL model, we use the same development set used by Snuba and GOGGLES for labeling. For training the upper bound model, we use the entire training set labels. The performance of each approach on hold-out test sets is reported in Table~\ref{table:end-to-end}.}

\revisenew{
First, GOGGLES outperforms Snuba by an average of 21\%, a similar number to the labeling performance improvement of 23\% GOGGLES has over Snuba (c.f. Table~\ref{tab:final-results}), and the end model performance of Snuba is worse than FSL. This is because the labels generated by Snuba ($59\%$) are only slightly better than random guessing, and having many extremely noisy labels can be more harmful than having fewer labels in training an end model. 
Second, GOGGLES outperforms the fine-tuned state-of-the-art FSL method (c.f. Section~\ref{sec:fsl}) by an average of 5\%, which is significant considering GOGGLES is only 7\% away from the upper bound. 
Third, not surprisingly, the more accurate the generated labels are, the more performance gain GOGGLES has over FSL (e.g., the improvements are more significant on CUB and Surface, which have higher labeling accuracies compared with other datasets).
}

\revisenew{
This experiment demonstrates the advantage GOGGLES has over FSL and data programming systems --- GOGGLES has the exact same inputs compared with FSL (both only have access to the pre-trained VGG-16 and the development set), and does not require dataset-specific labeling functions needed by data programming systems. 
}

\input{520-End2EndResults.tex}

%% file: 510-LabelingResults.tex
\begin{table*}[t]
\begin{tabular}{l|r|rr|rr|rrr}
\toprule
\multirow{2}{*}{Dataset} & \multicolumn{1}{c|}{GOGGLES}       & \multicolumn{2}{c|}{Data Programming} & \multicolumn{2}{c|}{Representation} & \multicolumn{3}{c}{Class Inference Baselines} \\
                         & \multicolumn{1}{c|}{(our results)} & Snorkel            & Snuba           & HoG                  & Logits                 & K-Means        & GMM         & Spectral       \\
\midrule
CUB                      & 97.83                             & 89.17                  & 58.83           & 62.93                 & 96.35                  & 98.67          & 97.62       & 72.08          \\
GTSRB                    & 70.51                             & -                  & 62.74           & 75.48                 & 64.77                  & 70.74          & 69.64       & 62.40          \\
Surface                  & 89.18                             & -                  & 57.86           & 85.82                 & 54.08                  & 69.08          & 69.14       & 60.82          \\
TB-Xray                  & 76.89                             & -                  & 59.47           & 69.13                 & 67.16                  & 76.33          & 76.70       & 75.00          \\
PN-Xray                  & 74.39                             & -                  & 55.50           & 53.11                 & 71.18                  & 50.66          & 68.66       & 75.90          \\
\midrule
Average                  & \textbf{81.76}                             & -                  & 58.88           & 69.30                 & 70.71                  & 73.09          & 76.35       & 69.24  \\
\bottomrule
\end{tabular}
\caption{\small{Evaluation of GOGGLES labeling accuracy on training set. The `-' symbol represents cases where evaluation was not possible. GOGGLES shows on average an improvement of 23\% over the state-of-the-art data programming system Snuba.}
}
\label{tab:final-results}
\vspace{-7mm}
\end{table*}

%% file: 520-End2EndResults.tex
\begin{table}[t!]
\begin{tabular}{l|r|rr|r|r}
\toprule
Dataset & FSL   & Snorkel & Snuba & GOGGLES & \begin{tabular}[l]{@{}l@{}}Upper\\ Bound\end{tabular} \\
\midrule
CUB     & 84.74 & 87.85   & 56.32 & 95.30   & \textcolor{gray}{98.44}                                \\
GTSRB   & 90.72 & -       & 70.11 & 91.54   & \textcolor{gray}{98.94}                                \\
Surface & 76.00 & -       & 51.67 & 83.33   & \textcolor{gray}{92.00}                                \\
TB-Xray & 66.42 & -       & 62.71 & 70.90   & \textcolor{gray}{82.09}                                \\
PN-Xray & 68.28 & -       & 62.19 & 69.06   & \textcolor{gray}{74.22}                                \\
\midrule
Average & 77.23 & -       & 60.60 & \textbf{82.03}   & \textcolor{gray}{89.14}                                \\
\bottomrule
\end{tabular}
\caption{\revisenew{\small{Comparison of end model accuracy on held-out test set. GOGGLES uses only 5 labeled instances per class but is only 7\% away from the supervised upper bound (in gray) which uses the ground-truth labels of the training set.}}}
\label{table:end-to-end}
\vspace{-10mm}
\end{table}

%% file: 600-RelatedWork.tex
\vspace{-2mm}
\section{Related Work}
\label{sec:relatedwork}

\stitle{ML Model Training with Insufficient Data.} 
Semi supervised learning techniques~\cite{zhu2005semi} combine labeled examples and unlabeled examples for model training; and active learning techniques aim at involving human labelers in a judicious way to minimize labeling cost~\cite{settles2012active}. Though semi-supervised learning and active learning can reduce the number of labeled examples required to obtain a competent model, they still need many labeled examples to start. 
\revisenew{Transfer learning ~\cite{pan2010survey} and few-shot learning techniques~\cite{DBLP:journals/corr/XianLSA17,fei2006one,DBLP:journals/corr/abs-1904-05046,2019ChenLKWH19} often use models trained on source tasks with many labeled examples to help with training models on new target tasks with limited labeled examples. Not surprisingly, these techniques often require users to select a source dataset or pre-trained model that is in a similar domain as the target task to achieve the best performance. In contrast, our proposal can incorporate several sources of information as affinity functions.}

\vspace{-1mm}
\stitle{Data Programming.} 
Data programming~\cite{ratner2016data}, and Snuba~\cite{varma2018snuba} and Snorkel~\cite{ratner2017snorkel} systems that implement the paradigm  were recently proposed in the data management community. Data programming focuses on reducing the human effort in training data labeling, and is the most relevant work to ours. Data programming ingests domain knowledge in the form of labeling functions. Each labeling function takes an unlabeled instance as input and outputs a label with better-than-random accuracy (or abstain). As we show in this paper, using data programming for image labeling tasks is particularly challenging, as it requires images to have associated metadata (e.g., text annotations or primitives), and a different set of labeling functions is required for every new dataset. In contrast, affinity coding and GOGGLES offer a domain-agnostic and automated approach for image labeling.

\vspace{-1mm}
\stitle{Other Related Work in Database Community.} Many  problems in database community share similar challenges to our work. In particular,data fusion/truth discovery~\cite{pochampally2014fusing,rekatsinas2017slimfast}, crowdsourcing~\cite{das2016towards}, and data cleaning~\cite{rekatsinas2017holoclean}, in one form or another, all need to reconcile information from multiple sources to reach one answer. While the information sources are assumed as input in these problems, labeling training data faces the challenge of lacking enough information sources. In fact, one primary contribution of GOGGLES is the affinity coding paradigm, where each unlabeled instance becomes an information source. 

%% file: 700-Conclusion.tex
\vspace{-2mm}
\section{Conclusion}
\label{sec:conclusion}
We proposed  affinity coding, a new paradigm that offers a domain-agnostic way of automated training data labeling. Affinity coding is based on the proposition that affinity scores of instance pairs belonging to the same class on average should be higher than those of instance pairs belonging to different classes, according to some affinity functions.  We build the GOGGLES system that implements the affinity coding paradigm for labeling image datasets. GOGGLES includes a novel set of affinity functions defined using the VGG-16 network, and a hierarchical generative model for class inference. GOGGLES is able to label images with high accuracy without any domain-specific input from users, except a very small development set, which is economical to obtain.

\vspace{-2mm}
\section{Acknowledgements}
\label{sec:acknowledgements}
We thank the SIGMOD'20 anonymous reviewers for their thoughtful and highly constructive feedback. This work was supported by NSF grants CNS-1704701 and IIS-1563816, and an Intel gift for ISTC-ARSA.